\documentclass[11pt,a4paper]{article}
\usepackage[hyperref]{emnlp2018}
\usepackage{times}
\usepackage{latexsym}

\usepackage{times}
\usepackage{latexsym}

\usepackage{url}

\usepackage{graphicx}
\usepackage{amsmath}
\usepackage{multirow}
\usepackage{booktabs}

\usepackage{url}
\usepackage{footmisc}

\aclfinalcopy 

\title{An Edit-centric Approach for Wikipedia Article Quality Assessment}

\author{
Edison Marrese-Taylor\textsuperscript{1}, Pablo Loyola\textsuperscript{2}, Yutaka Matsuo \textsuperscript{1}\\
Graduate School of Engineering, The University of Tokyo, Japan\textsuperscript{1} \\
{\tt \{emarrese, matsuo\}@weblab.t-utokyo.ac.jp}\\
IBM Research, Tokyo, Japan\textsuperscript{2}\\ 
{\tt e57095@jp.ibm.com}}

\date{}

\begin{document}
\maketitle

\begin{abstract}
We propose an edit-centric approach to
assess Wikipedia article quality as a complementary
alternative to current full document-based techniques. Our model consists
of a main classifier equipped with an auxiliary 
generative module which, for a given edit, jointly provides an 
estimation of its quality and generates a description in natural
language. 
We performed an empirical study to assess the feasibility of the proposed model
and its cost-effectiveness in terms of data and quality requirements.
\end{abstract}

\section{Introduction}





Wikipedia is arguably the world's most famous example of crowd-sourcing involving natural language. Given its open-edit nature, article often end up containing passages that can be regarded as noisy. These may be the indirect result of benign edits that do not meet certain standards, or a more direct consequence of vandalism attacks. In this context, assessing the quality of the large and heterogeneous stream of contributions is critical for maintaining Wikipedia's reputation and credibility. To that end, the WikiMedia Foundation has deployed a tool named ORES \cite{halfaker2015artificial} to help monitor article quality, which treats quality assessment as a supervised multi-class classification problem. This tool is static, works at the document-level, and is based on a set of predefined hand-crafted features \cite{warncke-wang_success_2015}.

While the ORES approach seems to work effectively, considering the whole document could have negative repercussions. As seen on Figure \ref{fig:ml}, article length naturally increases over time (see Appendix \ref{subsed:wiki_growth} for additional examples), which could lead to scalability issues and harm predictive performance, as compressing a large amount of content into hand crafted features could diminish their discriminative power. We conducted an exploratory analysis using a state-of-the-art \cite{dang_quality_2016} document-level approach, finding that there is a clear negative relationship between document length and model accuracy (see details on Appendix \ref{subsec:doc_length}). 

\begin{figure}
    \centering
    \includegraphics[width=0.9\linewidth, height=2.5cm]{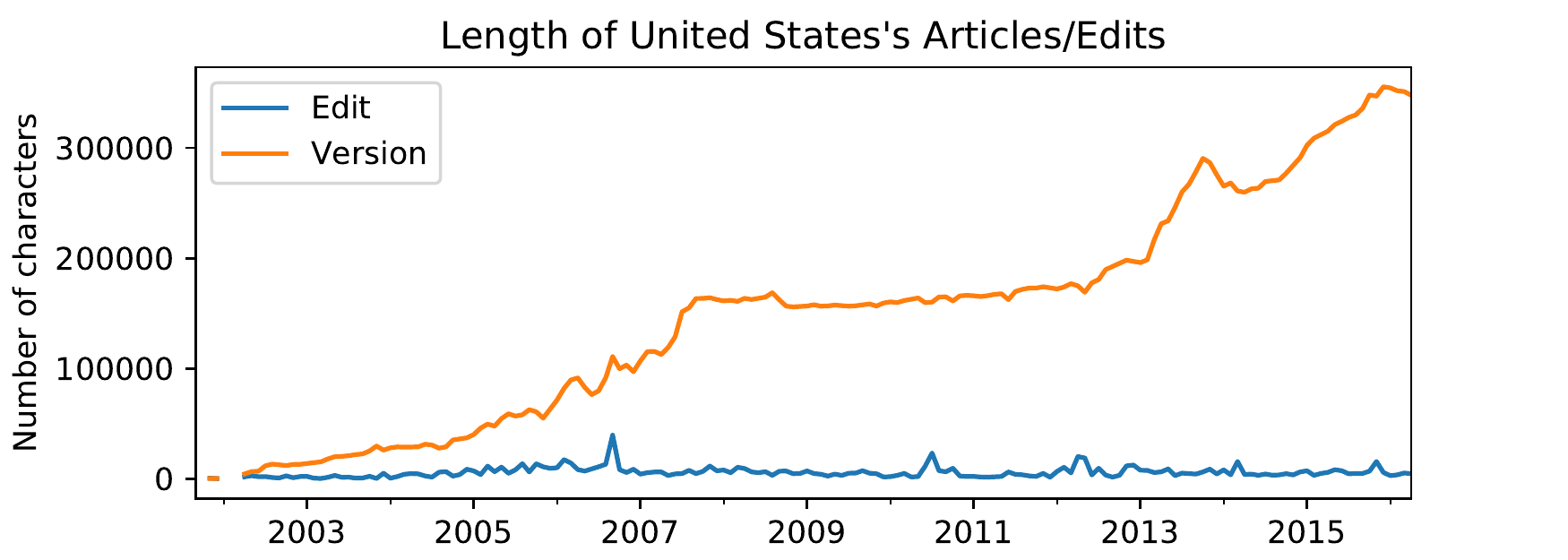}
    \caption{Monthly average of the article/edits length (in number of characters) for the United States article.}
    \label{fig:ml}
\end{figure}

In light of this, we are interested in exploring a complementary alternative for assessing article quality in Wikipedia. We propose a model that receives as input only the edit, computed as the difference between two consecutive article versions associated to a contribution, and returns a measure of article quality. As seen on Figure \ref{fig:ml}, edit lengths exhibit a more stable distribution over time.

Moreover, as edits are usually accompanied by a short description  which clarifies their purpose and helps with the reviewing process \citep{Guzman:2014:SAC:2597073.2597118}, we explore whether this information could help improve quality assessment by also proposing a model that jointly predicts the quality of a given edit and generates a description of it in natural language. Our hypothesis is that while both tasks may not be completely aligned, the quality aspect could be benefited by accounting for the dual nature of the edit representation.

We performed an empirical study on a set of Wikipedia pages and their edit history, evaluating the feasibility of the approach. Our results show that our edit-level model offers competitive results, benefiting from the proposed auxiliary task. In addition to requiring less content as input, we believe our model offers a more natural approach by focusing on the actual parts of the documents that were modified, ultimately allowing us to transition from a static, document-based approach, to an edit-based approach for quality assessment.

\section{Related Work}



In terms of quality assessment, the pioneer work of \citet{hu2007measuring} used the interaction between articles and their editors to estimate quality, proposed as a classification task. Later, \citet{kittur2008harnessing} studied how the number of editors and their coordination affects article quality, while \citet{blumenstock2008size} proposed to use word count for measuring article quality. 

\citet{warncke-wang_tell_2013,warncke-wang_success_2015} took the classification approach and characterized an article version with several hand-crafted features, training a SVM-based model whose updated version was deployed into the ORES system. More recent work has experimented with models based on representation learning, such as \citet{dang_quality_2016} who used a doc2vec-based approach, and \citet{shen_hybrid_2017} who trained an RNN to encode the article content. While all these models are inherently static, as they model the content of a version, the work of \citet{zhang_history-based_2018} is, to the best of our knowledge, the only one to propose a history-based approach.

On the other hand, \citet{su_liu_2015} tackled the quality problem by using a psycho-lexical resource, while \citet{kieseletal_2017} aimed at automatically detecting vandalism utilizing change information as a primary input. \citet{gandon_etal_2016} also validated the importance of the editing history of Wikipedia pages as a source of information.

In addition to quality assessment, our work is also related to generative modeling on Wikipedia. Recent work includes approaches based on autoencoders, such as \citet{chisholm2017learning}, who generate short biographies, and \citet{yin2018learning} who directly learn diff representations. Other works include the approach by \citet{zhang2017wikum} which summarizes the discussion surrounding a change in the content, and by \citet{boyd_using_2018} who utilizes Wikipedia edits to augment a small dataset for grammatical error correction in German.


\section{Proposed Approach}
Our goal is to model the quality assessment task on Wikipedia articles from a dynamic perspective. Let $v_1, \dots, v_T$ be the sequence of the time-sorted $T$ revisions of a given article in Wikipedia. Given a pair of consecutive revisions $(v_{t-1}, v_{t})$, an \textit{edit} $e_t = \Delta_{t-1}^{t}(v)$ is the result of applying the Unix \textit{diff} tool over the \textit{wikitext}\footnote{https://www.mediawiki.org/wiki/Wikitext} contents of the revision pair, allowing us to recover the added and deleted lines on each edition. 

Due to the line-based approach of the Unix \textit{diff} tool, small changes in \textit{wikitext} may lead to big chunks (or hunks) of differences in the resulting \textit{diff} file. Moreover, as changes usually occur at the sentence level, these chunks can contain a considerable amount of duplicated information. To more accurately isolate the introduced change, we segment the added and removed lines on each hunk into sentences, and eliminate the ones appearing both in the added and removed lines. Whenever multiple sentences have been modified, we use string matching techniques to identify the before-after pairs. After this process, $e_t$ can be characterized with a set of before-after sentence pairs ${(s_{ti}^{-}, s_{ti}^{+})}$, where $s_{ti}^{+}$ is an empty string in case of full deletion, and vice-versa. 

Similarly to \citet{yin2018learning}, to obtain a fine-grained characterization of the \textit{edit}, we tokenize each sentence and then use a standard \textit{diff} algorithm to compare each sequence pair. We thus obtain an alignment for each sentence pair, which in turn allows us to identify the tokens that have been added, removed, or remained unchanged. For each case, we build an \textit{edit-sentence} based on the alignment, containing added, deleted and unchanged tokens, where the nature of each is characterized with the token-level labels $+$, $-$ and $=$, respectively.

For a given \textit{edit} $e_t$ we generate an \textit{edit} representation based on the contents of the associated \textit{diff}, and then use it to predict the quality of the article in that time. We follow previous work and treat quality assessment as a multi-class classification task. We consider a training corpus with $T$ edits $e_t$, $1 \leq t \leq T$. 


Our quality assessment model encodes the input \textit{edit-sentence} using a BiLSTM. Concretely, we use a token embedding matrix $E_T$ to represent the input tokens, and another embedding matrix $E_L$ to represent the token-level labels. For a given embedded token sequence $X_t$ and embedded label sequence $L_t$, we concatenate the vectors for each position and feed them into the BiLSTM to capture context. We later use a pooling layer to obtain a fixed-length edit representation.


\subsection{Incorporating Edit Message Information}


When a user submits an \textit{edit}, she can add a short message describing or summarizing it. We are interested in studying the how these  messages can be used as an additional source to support quality assessment task. A natural, straightforward way to incorporate the message into our proposal is to encode it into a feature vector using another BiLSTM with pooling, and combine this with the features learned from the \textit{edit}. 

Furthermore, we note that the availability of an edit message actually converts an \textit{edit} into a dual-nature entity. In that sense, we would like to study whether the messages are representative constructs of the actual \textit{edits}, and how this relation, if it exists, could impact the quality assessment task. One way to achieve this is by learning a mapping between \textit{edits} and their messages.  

Therefore, we propose to incorporate the edit messages by adding an auxiliary task that consists of generating a natural language description of a given \textit{edit}. The idea is to jointly train the classification and auxiliary tasks to see if the performance on quality assessment improves. Our hypothesis is that while both tasks are not naturally aligned, the quality aspect could benefit by accounting for the dual nature of the \textit{edit} representation.
 
Our proposed generative auxiliary task is modeled using a sequence-to-sequence \cite{sutskever2014sequence,cho_learning_2014} with global attention \cite{bahdanau2014neural, luong-pham-manning:2015:EMNLP} approach, sharing the encoder with the classifier. During inference, we use beam search and let the decoder run for a fixed maximum number of steps, or until the special end-of-sentence token is generated. This task is combined with our main classification task using a linear combination of their losses, where parameter $\lambda$ weights the importance of the classification loss.


\section{Empirical Study}


 We collected historical dumps from Wikipedia, choosing some of the most edited articles for both the English and German languages. Wikipedia dumps contain every version of a given page in \textit{wikitext}, along with metadata for every \textit{edit}. To obtain the content associated to each $\Delta_{t-1}^{t}(v)$, we sorted the extracted \textit{edits} chronologically and computed the \textit{diff} of each pair of consecutive versions using the Unix diff tool. We ignore \textit{edits} with no accompanying message. For English sentence splitting we used the automatic approach by \citet{kiss2006}, and Somajo \cite{somajo2016} for German. The quality labels are obtained using the ORES API, which gives us a probability distribution over the quality labels for each revision that we use as a silver standard. We randomize and then split each dataset using a 70/10/20 ratio.

For comparison, we also consider the \textit{Wikiclass} dataset built by \citet{warncke-wang_success_2015}, which consists of 30K revisions of random Wikipedia articles paired with their manually-annotated quality classes. To use this dataset with our models, we identified and downloaded the page revision immediately preceding each example using the Wikipedia API, to later apply the Unix \textit{diff} tool and obtain the \textit{edits}. We use the train/test splits provided and 20\% of the training set as a validation. Other similar datasets are not suitable for us as they do not include the revision ids which we require in order to obtain the edits.


\subsection{Experimental Setting}

On our collected datasets, the classification models are trained using the Kullback-Leibler divergence as the loss function ---which in our preliminary experiments worked better than using the derived hard labels with cross entropy--- while for the \textit{Wikiclass} dataset we used the cross entropy with the gold standard. In both cases we used accuracy on the validation set for hyper-parameter tuning and evaluation, and also measured macro-averaged F1-Score. For the models with the auxiliary task, we also evaluate our generated descriptions with sentence-level BLEU-4 \cite{papineni-EtAl:2002:ACL}.


\subsection{Results}

We firstly conducted an ablation study to identify the model components that have greater impact on the performance. We compare our \textit{edit-sentence} encoder with a regular encoding mechanism, where the tokens from $s_{ti}^{-}$ and $s_{ti}^{+}$ are concatenated (separated with a special marker token), and with a version that ignores the token-level label embeddings. 

\begin{table}[h!]
    \centering
    \scriptsize
    \begin{tabular}{c c c c }
        \toprule
        \textbf{Model} & \textbf{F1} & \textbf{Acc} & \textbf{BLEU}\\
        \midrule
        Regular & 0.47 & 0.74 & - \\
        + \textit{edit-sentence} & 0.56 & \textbf{0.80} & - \\
        + \textit{diff} tags  & 0.62 & 0.78 & - \\
        \midrule
        + Generation $\lambda=0.2$ & 0.28 & 0.61 & 0.25 \\
        + Generation $\lambda=0.5$ & 0.33 & 0.68 & 0.24 \\
        + Generation $\lambda=0.8$ & 0.41 & 0.77 & 0.25 \\
        + Generation $\lambda=0.9$ & \textbf{0.65} & 0.77 & 0.22 \\
        \midrule
        Only Generation ($\lambda=0$) & - & - & 0.23 \\
        \bottomrule
    \end{tabular}
    \caption{Impact of the parameters on validation performance for the WWII article history.}
    \label{table:ablation}
\end{table}

As seen on Table \ref{table:ablation}, when compared against the regular encoder, utilizing our \textit{edit-sentence} approach with token-level labels leads to a higher F1-Score and accuracy, showing the effectiveness of our proposed \textit{edit} encoder. These results also shed some light on the trade-off between tasks for different values of $\lambda$. We see that although a higher value tends to give better classification performance both in terms of F1-Score and accuracy, it is also possible to see that there is a sweet-spot that allows the classification to benefit from learning an \textit{edit}-message mapping, supporting our hypothesis. Moreover, this comes at a negligible variation in terms of BLEU scores, as seen when we compare against a pure message-generation task (Only Generation on the table).

On the other hand, when we tested the alternative mechanism to combine the edit and message information simply combining their  representations and feeding them to the classifier, we obtained no performance improvements. This again supports our choice to model the \textit{edit}-message mapping for the benefit of quality assessment. 

Since we discarded \textit{edits} that were not accompanied by messages during pre-processing, it is difficult to assess the impact that the absence of these messages may have on quality assessment. In those cases, we believe our model with the auxiliary generative task could be used as a drop-in replacement and thus help content reviewers. 

\begin{table}[h]
    \centering
    \scriptsize
    \begin{tabular}{c c c c c c c c}
        \toprule
        \multicolumn{2}{c}{\multirow{2}{*}{\textbf{{\shortstack{Dataset\\Model}}}}} & \multicolumn{3}{c}{\textbf{Validation}} & \multicolumn{3}{c}{\textbf{Test}}\\
        \cmidrule(lr){3-5} \cmidrule(lr){6-8}
        & & \textbf{F1} & \textbf{Acc} & \textbf{BL} & \textbf{F1} & \textbf{Acc} & \textbf{BL} \\
        \midrule
        \multirow{2}{*}{\shortstack{Barack\\Obama}}
                & C & 0.50 & 0.92 & - & 0.62 & \textbf{0.91} & - \\
                & C+G & 0.57 & 0.89 & 0.21 &\textbf{0.66}& 0.88 & 0.20 \\
        \midrule
        \multirow{2}{*}{\shortstack{Donald\\Trump}}
                & C & 0.69 & 0.79 & - & \textbf{0.47}& \textbf{0.78} & - \\
                & C+G & 0.69 & 0.76 & 0.22 & \textbf{0.47} & 0.77 & 0.20 \\
        \midrule
        \multirow{2}{*}{\shortstack{Guns n'\\Roses}} 
                & C   & 0.28 & 0.86 & - & 0.18 & \textbf{0.84} & -  \\
                & C+G & 0.31 & 0.79 & 0.23 & \textbf{0.30} & 0.81 &  \\
        \midrule
        \multirow{2}{*}{Xbox 360} 
                & C   & 0.22 & 0.60 & -    & 0.30 & 0.61 & -  \\
                & C+G & 0.37 & 0.62 & 0.34 & \textbf{0.32} & \textbf{0.63} & 0.31 \\
        \midrule
        \multirow{2}{*}{Chicago} 
                & C   & 0.36 & 0.71 & - & 0.38 & \textbf{0.72} & -  \\
                & C+G & 0.44 & 0.70 & 0.30 & \textbf{0.39} & 0.71 & 0.29 \\
        \midrule
        \multirow{2}{*}{\shortstack{Pink\\Floyd}} 
                &   C & 0.43 & 0.79 & -    & 0.35 & 0.80 & -  \\
                & C+G & 0.46 & 0.80 & 0.34 & \textbf{0.37} & \textbf{0.80} & 0.35 \\
        \midrule
        \multirow{2}{*}{\shortstack{Manchester\\United F.}} 
                & C   & 0.15 & 0.24 & -    & 0.17 & 0.72 & -  \\
                & C+G & 0.29 & 0.79 & 0.43 & \textbf{0.39} & \textbf{0.77} & 0.43 \\
        \midrule
        \midrule                                    
        \multirow{2}{*}{Deutschland}
                & C & 0.19 & 0.31 & - & 0.12 & 0.31 & - \\
                & C+G & 0.22 & 0.38 & 0.31 & \textbf{0.17 }& \textbf{0.33} & 0.36\\
        \midrule
        \multirow{2}{*}{\shortstack{Zweiter\\Weltkrieg}} 
                & C   & 0.26 & 0.29 & - & 0.15 & \textbf{0.30} & -  \\
                & C+G & 0.29 & 0.32 & 0.35 & \textbf{0.18} & \textbf{0.30} & 0.28 \\
        \bottomrule
    \end{tabular}
    \caption{Summary of our results. C indicates models that only perform classification, and C+G multi-task models. BL is short for BLEU-4.}
    \label{table:results}
\end{table}

Table \ref{table:results} summarizes our best results on each selected article. We see how the addition of the generative task can improve the classification performance for both considered languages. In terms of the task trade-off, controlled with parameter $\lambda$, we empirically found that higher values tend to work better for datasets with more examples. 

Regarding the \textit{Wikiclass} dataset, we compared our model against a state-of-the-art document-level approach \cite{dang_quality_2016, shen_hybrid_2017} based on on doc2vec \cite{le_distributed_2014}. In this scenario, our model obtains an accuracy of 0.40 on the test set, while the document level approach reaches 0.42. While the document level approach performed slightly better, our model is able to obtain a reasonable performance in a more efficient manner as it requires an input that averages only 2K characters (the edits), which contrasts to the average 12K characters in the documents. It is worth mentioning that the performance of the document-level approach reported by \citet{dang_quality_2016} significantly differs from the value reported here. By looking at their implementation\footnote{\url{github.com/vinhqdang/doc2vec\_dnn\_wikipedia}} we note that this value is obtained when also using the test documents to train. 






\section{Conclusion and  Future work}

In this work we proposed a new perspective to the problem of quality assessment in Wikipedia articles, taking as central element the dynamic nature of the \textit{edits}. Our results support our hypothesis and show the feasibly of the approach. We believe the  temporal view on the problem that the proposed approach provides could open the door to incorporating behavioral aspects into the quality estimation, such as user traces and reverting activity, which are also critical to limit the amount of noise and ensure the reliability of Wikipedia. 



\clearpage
\newpage

\bibliography{bibliography}
\bibliographystyle{acl_natbib}

\clearpage
\newpage

\appendix

\section{Supplemental Material}
\label{sec:supplemental}

\subsection{Wikipedia Article Growth}
\label{subsed:wiki_growth}

\begin{figure}[h!]
    \centering
    \includegraphics[width=0.9\linewidth]{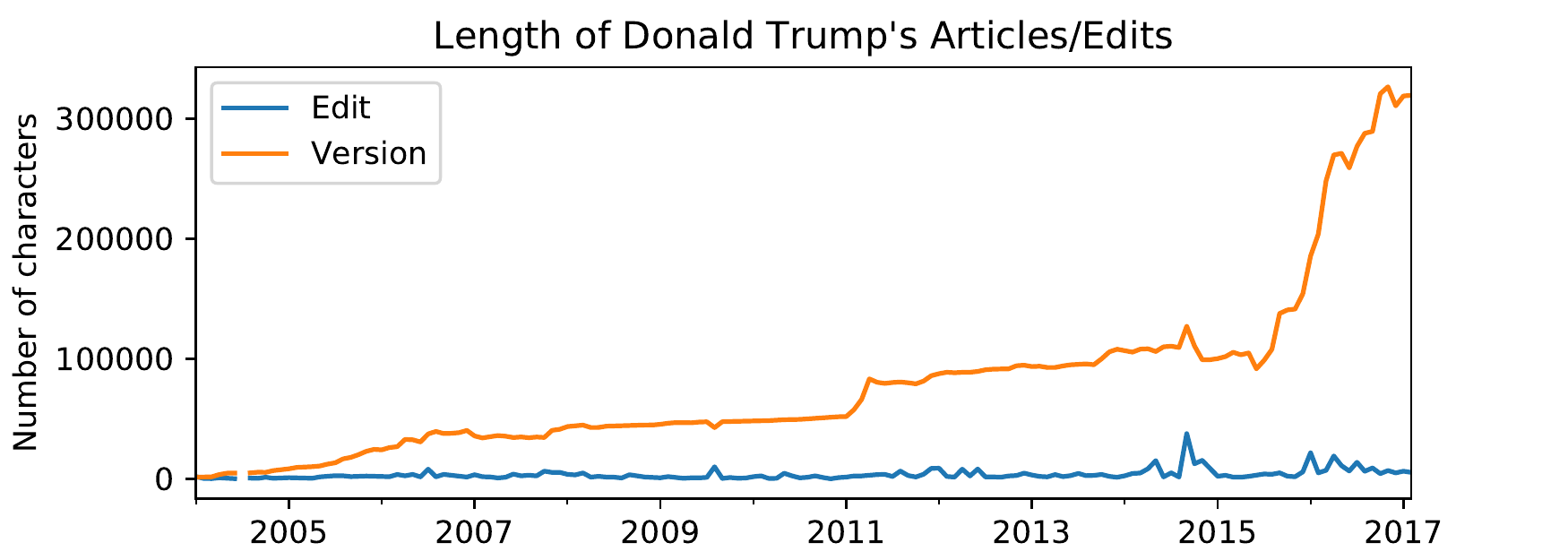}
    \includegraphics[width=0.9\linewidth]{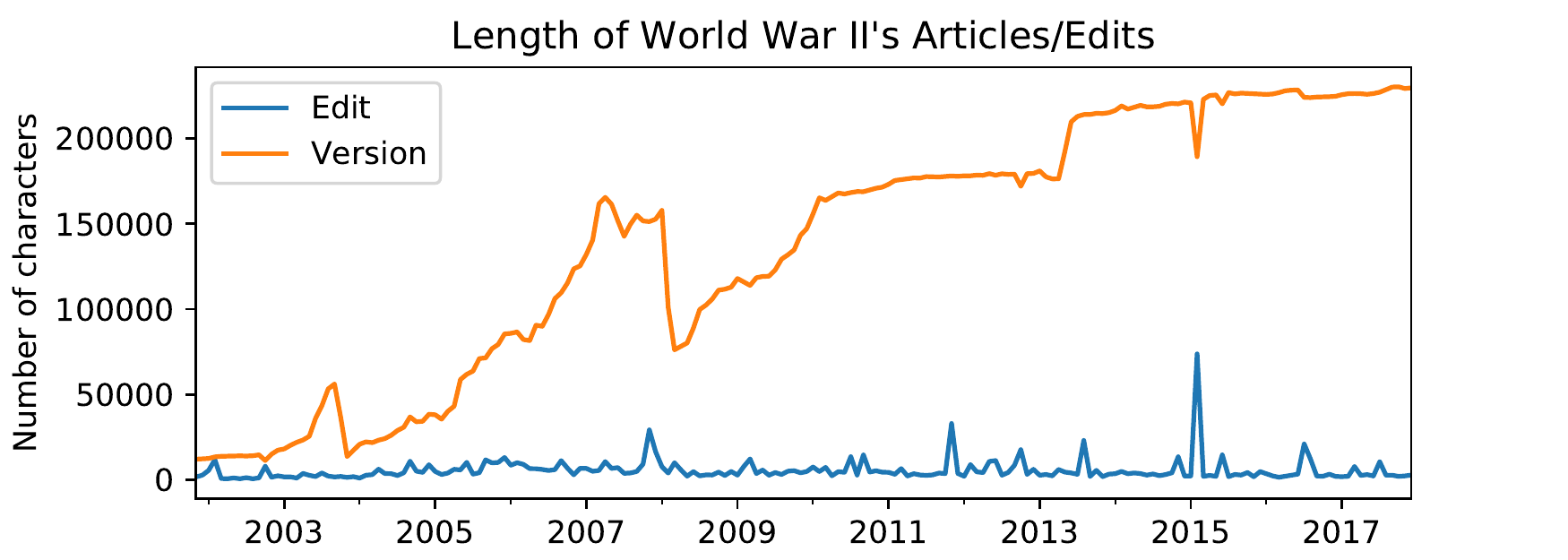}
    \includegraphics[width=0.9\linewidth]{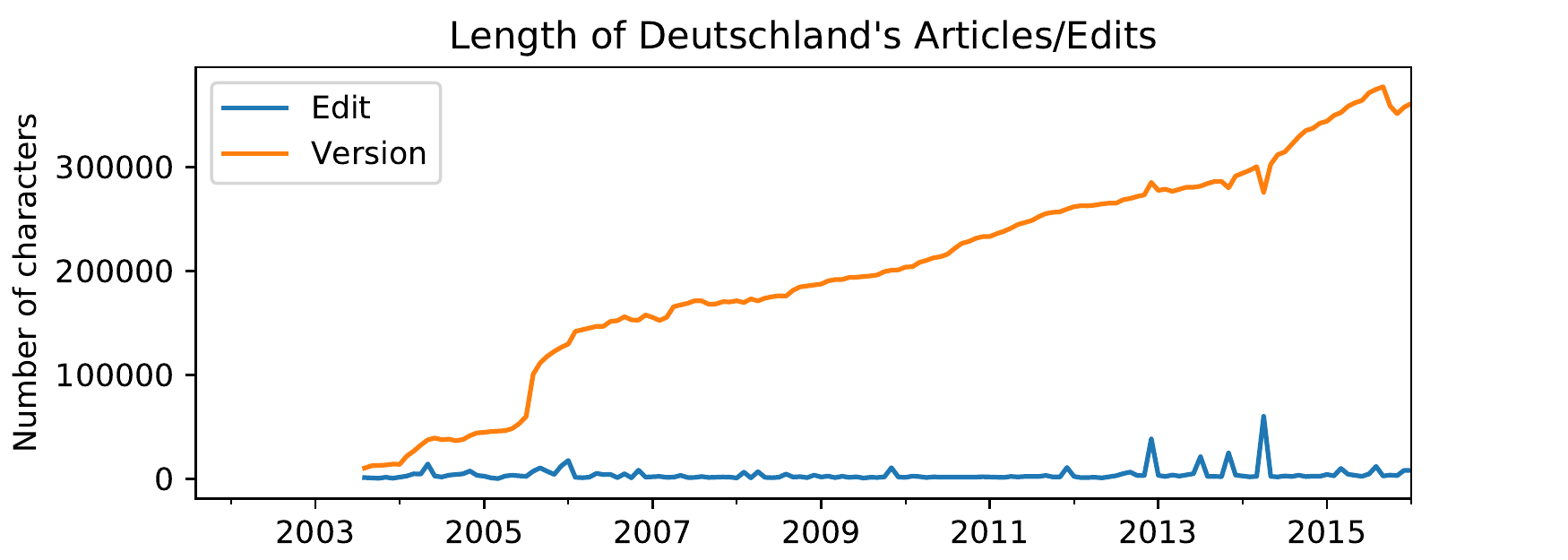}
    \includegraphics[width=0.9\linewidth]{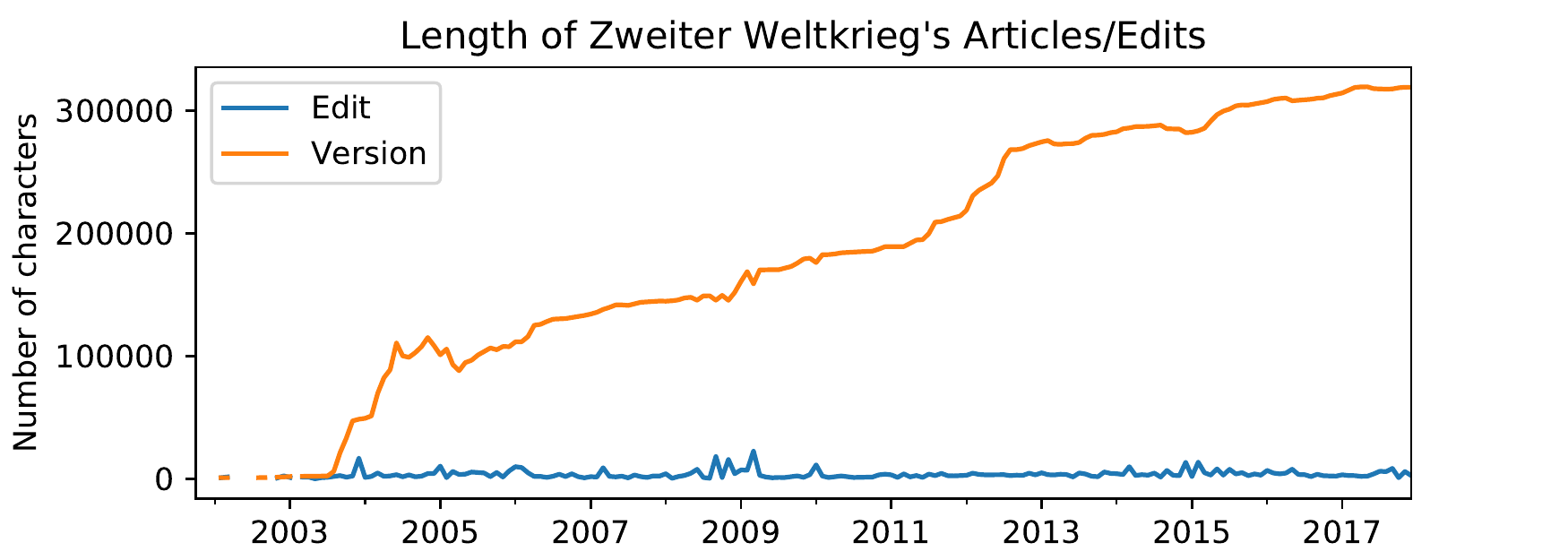}
    \caption{Monthly average of the article/edits length (in number of characters) for different Wikipedia articles.}
    \label{fig:wiki_growth}
\end{figure}

Figure \ref{fig:wiki_growth} shows how the monthly average of the article/edits length (in number of characters) varies over time for 4 different Wikipedia pages in two different languages: Donald Trump and World War II in the English Wikipedia, and Deutschland (Germany) and Zweiter Weltkrieg (World War II) in the German Wikipedia. It is possible to see that the average Wikipedia article size has been increasing dramatically over the years, and that this tendency seems to generalize across languages. In contrast, we also see that the average size of the edits applied remain relatively constant and that these are comparatively short. 

\newpage
\subsection{Impact of document length on performance}
\label{subsec:doc_length}
We implement the approach by \citet{dang_quality_2016} directly based on their code release, available on GitHub\footnote{\url{https://github.com/vinhqdang/doc2vec_dnn_wikipedia}}. Their implementation uses the test documents when training the doc2vec model, which we consider inadequate. Instead we train only using the documents in the training split. The results for the original setting, although not reported here, are similar.

\begin{figure}[h]
    \centering
    \includegraphics[width=0.9\linewidth]{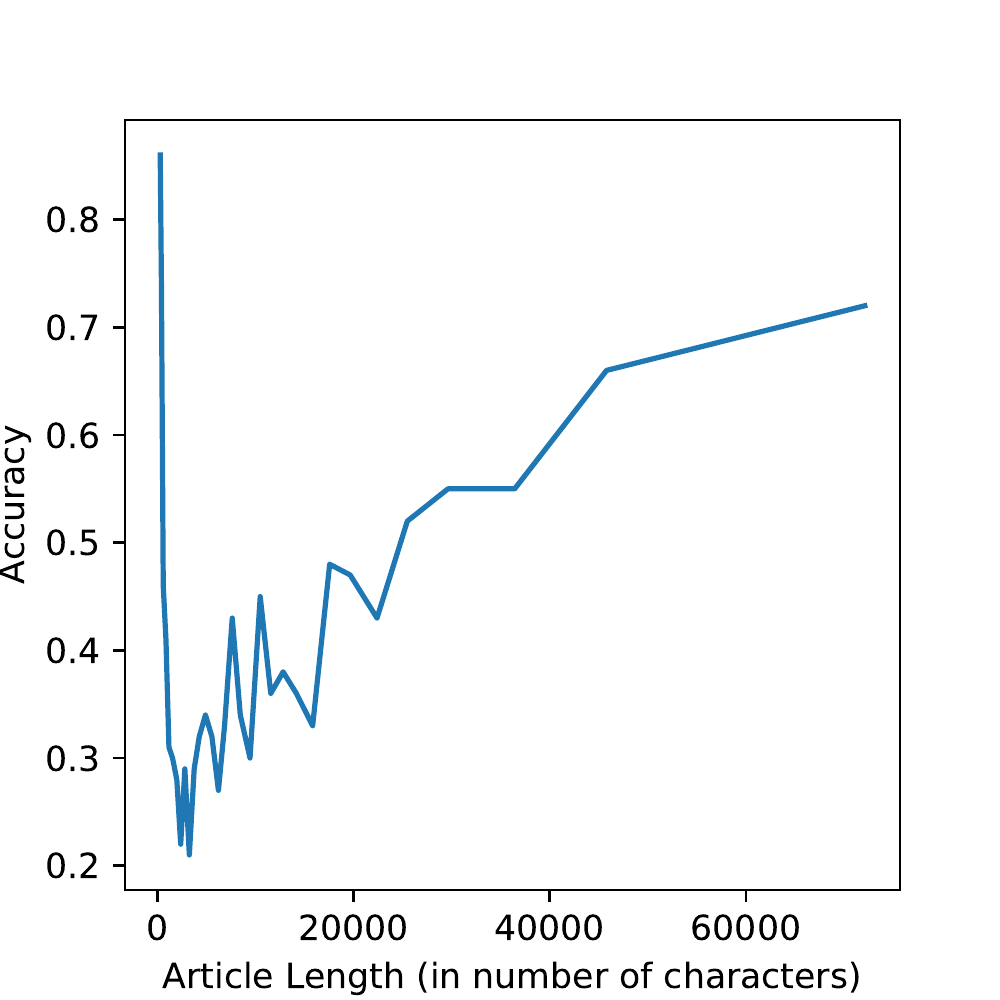}
    \caption{Performance of the doc2vec-based approach on the test split of the \textit{Wikiclass} dataset, for different input lengths.}
    \label{fig:len_perf}
\end{figure}

Table \ref{fig:len_perf} shows how the performance the doc2vec-based approach on the test split of the \textit{Wikiclass} dataset for different document lengths, in characters. This model is regarded as the state-of-the-art model on this dataset that does not require hand-crafted features.

\end{document}